# A Comparative Study on Deep-Learning Methods for Dense Image Matching of Multi-angle and Multi-date Remote Sensing Stereo Images

Hessah Albanwan[1,2] and Rongjun Qin[1, 2, 3,4,*]

[1] Geospatial Data Analytics Laboratory, The Ohio State University, 218B Bolz Hall, 2036 Neil Avenue, Columbus, OH 43210, USA
[2] Department of Civil, Environmental and Geodetic Engineering, The Ohio State University, 218B Bolz Hall, 2036 Neil Avenue, Columbus, OH 43210, USA
[3] Department of Electrical and Computer Engineering, The Ohio State University, 205 Dreese Lab, 2036 Neil Avenue, Columbus, OH 43210, USA
[4] Translational Data Analytics Institute, The Ohio State University, USA
Email: Albanwan.1@osu.edu, Qin.324@osu.edu
*Corresponding author

***Abstract***: Deep learning (DL) stereo matching methods gained great attention in remote sensing satellite datasets. However, most of these existing studies conclude assessments based only on a few/single stereo images lacking a systematic evaluation on how robust DL methods are on satellite stereo images with varying radiometric and geometric configurations. This paper provides an evaluation of four DL stereo matching methods through hundreds of multi-date multi-site satellite stereo pairs with varying geometric configurations, against the traditional well-practiced Census-SGM (Semi-global matching), to comprehensively understand their accuracy, robustness, generalization capabilities, and their practical potential. The DL methods include a learning-based cost metric through convolutional neural networks (MC-CNN) followed by SGM, and three end-to-end (E2E) learning models using Geometry and Context Network (GCNet), Pyramid Stereo Matching Network (PSMNet), and LEAStereo. Our experiments show that E2E algorithms can achieve upper limits of geometric accuracies, while may not generalize well for unseen data. The learning-based cost metric and Census-SGM are rather robust and can consistently achieve acceptable results. All DL algorithms are robust to geometric configurations of stereo pairs and are less sensitive in comparison to the Census-SGM, while learning-based cost metrics can generalize on satellite images when trained on different datasets (airborne or ground-view).

## I. INTRODUCTION

Stereo dense image matching (DIM) has been an active area of study over the years, as it offers a cost-effective and efficient approach to generate digital surface models (DSM) for applications such as 3D modeling, forestry mapping, and change detection (Albanwan et al., 2020; Albanwan & Qin, 2020; Furukawa & Hernández, 2015; Navarro et al., 2018; Nebiker et al., 2014; Tian et al., 2014; Wang et al., 2017). This is especially relevant when these applications are fueled by satellite-based 3D reconstructions due to their wide data coverage and consistent data collection over time (Albanwan & Qin, 2020; Qin, 2017). However, it has been noted (Brown et al., 2003; Qin, 2019; Seitz et al., 2006) that the quality of the DSM not only depends on a specific DIM algorithm, but also on various factors of stereo pairs including: 1) sensor and image characteristics (e.g., spatial and radiometric resolutions), 2) acquisition conditions including the atmosphere and position and orientation of the sun, camera, and objects, 3) scene structure and texture across different geographical regions, for instance, different land patterns including urban, suburban, or forest areas, water surfaces, and parallaxes formed by stereo pairs, etc. Although there have been studies evaluating the achievable quality of these satellite-based reconstructions using different DIM algorithms, most of them only take one or a few pairs for quality assessment, thus the resulting conclusions are often not comprehensive to cover data with various stereo configurations appearing in practice.
A typical stereo DIM algorithm performs disparity estimation on rectified stereo images (i.e., epipolar images), which generally follows several key steps, including cost matching, aggregation, disparity optimization, and filtering (Scharstein et al., 2001). Cost matching is the main step for measuring feature similarities for disparity computation. For the last decade, Census has been known as the classical cost matching metric for stereo DIM problems and has been studied intensively due to its robustness to radiometric differences, at the same time it can be computed efficiently (Chen et al., 2019; Ma et al., 2013; Xia et al., 2018; Zabih & Woodfill, 1994). On the other hand, recently developed deep learning (DL) algorithms have shown a promising performance that is assumed to surpass classical cost metrics like Census (Chen et al., 2019; Hamid et al., 2020). There are mainly two ways that DL algorithms are implemented in stereo DIM, for example, 1) Learning-based cost metric and 2) end-to-end (E2E) learning. The learning-based cost metric learns similarity from image patches given examples of similar and dissimilar patches, such methods were first introduced by (Žbontar & LeCun, 2015) through convolutional neural networks (MC-CNN) as learnable feature extractors. Like Census, such patch-based cost metrics do not handle texture-less regions well, and often require multistage optimization after cost computation, followed by cost aggregation for regularization and smoothness of the disparity map (Bobick & Intille, 1999; Hirschmuller, 2005, 2008; Kolmogorov & Zabih, 2001). There are plenty of cost aggregation methods ranging from local, semi-global, and global approaches. A well-known example is the semi-global matching (SGM) algorithm (Hirschmuller, 2005), which is known to effectively leverage both accuracy and speed well, thus nowadays it is used as a standard cost aggregation approach. However, even with cost aggregation, studies found that MC-CNN still suffers from poor performance in ill-posed regions with occlusion, high-reflective surfaces, lack of texture, and repetitive patterns (Chen et al., 2019; Ma et al., 2013; Xia et al., 2018; Zabih & Woodfill, 1994) demanding further postprocessing and refinement. Alternatively, E2E learning methods directly generate disparity from stereo pair rectified images without additional optimization steps. They have become a popular line of stereo DIM algorithms because they can directly predict highly accurate disparity maps through learning from geometry and context (e.g., cues such as shading, illumination, objects, etc.) rather than low-level features (Chang & Chen, 2018; Cheng et al., 2020; Gu et al., 2020; Kendall et al., 2017; Xu & Zhang, 2020; Zhang, Prisacariu, et al., 2019). Examples of this type of work are Geometry and Context Network (GCNet), pyramid stereo matching networks (PSMNet), and LEAStereo, which due to their high accessibility and performances (i.e., winning the best rank in the KITTI 2012 and 2015 leaderboards (Chang & Chen, 2018; Cheng et al., 2020; Kendall et al., 2017), have become popular in the field.

However, DL algorithms often suffer from generalization problems, at the same time, face challenges to process large-volume and large-format remote sensing images, limiting their practical values (Pang et al., 2018; Song et al., 2021; Zhang, Qi, et al., 2019). The degree of generalization may vary with both DL models and tasks. For instance, studies indicate that most E2E approaches enjoy deep feature representations, which however, may encode scene-specific information (Pang et al., 2018; Song et al., 2021; Zhang, Qi, et al., 2019), thus may poorly perform on unseen scenes or data (Song et al., 2021). Most of the existing DL stereo DIM algorithms are trained using Computer Vision (CV) benchmark datasets such as KITTI or Middlebury, etc. (Geiger et al., 2012; Hamid et al., 2020), which mainly consist of images of everyday scenes. These images are distinctively different from satellite images in terms of scene content, view perspectives, and object granularity, thus DL models trained from these CV datasets may not perform well when applied to satellite datasets. However, acquiring a large number of high-quality satellite datasets for training is a challenge mainly due to its high cost. Additionally, the scene content of the earth's surface is extremely diverse across the entire Globe, thus it is difficult to create a reprehensive dataset by only using data from a few regions, although these regions can be as large as an entire city. A few remote sensing benchmarks such as the IARPA (The Intelligence Advanced Research Projects Activity) Multiview stereo 3D mapping challenge (Bosch et al., 2016) and the 2019 Institute of Electrical and Electronics Engineers (IEEE) Geoscience and Remote Sensing Society (GRSS) Data Fusion Contest (DFC) (Le Saux, 2019) provide the ground-truth dataset in the form of LiDAR point clouds and raster DSMs, which must be post-processed and converted to the ground-truth disparity for training. Unfortunately, this step may generate undesired errors in the training data due to geometric errors in the orientation parameters, inconsistencies in the temporal and spatial resolutions, or other operations such as triangulation, projection, and interpolation (Bosch et al., 2016; Cournet et al., 2020; Patil et al., 2019; Wu et al., 2021).

In addition, a significantly under-evaluated criterion for DIM algorithms on satellite images is their robustness to varying stereo configurations such as sun illuminations, intersection angles, different sensors, etc. A robust DIM algorithm pertaining to these factors is extremely important because this will allow us to make full use of the already limited satellite images (as compared to everyday images). Ultimately, we want these algorithms to be agnostic and less selective to stereo configurations of data. It has been reported that the result of a typical stereo algorithm, i.e., Census with SGM, directly correlates with geometrical acquisition factors such as sun angle difference and intersection angle (Qin, 2019). Unfortunately, the same analysis has not been covered for DL algorithms. Since DL methods (especially E2E ones) can learn context, it is of interest to understand their performance under varying stereo configurations.

In this paper, we aim to comprehensively explore the limitations and strengths of recent stereo DIM algorithms for satellite datasets. We consider representative DIM algorithms of three main categories: 1) traditional approaches (e.g., Census cost metric with SGM), 2) deep learning-based cost metrics (e.g., MC-CNN cost matching metric with SGM), and 3) three E2E learning methods (e.g., GCNet, PSMNet, and LEAStereo). Since "Census+SGM" has been well studied and widely used in satellite stereo-photogrammetry (Qin, 2019), it serves as a baseline method in this study. Deep learning-based cost metric is regarded as a simpler task than E2E learning for DIM because it learns patch-level similarity as a binary classification problem (similar or not similar). The very one algorithm in this category is MC-CNN, which applies a Siamese network to learn the similarities (Žbontar & LeCun, 2015). Despite the very many E2E methods in the CV community (Laga, 2019), we choose three State-of-the-art (SOTA) methods that are frequently used by the community, and well-performed in the leader board (Chang & Chen, 2018; Cheng et al., 2020; Kendall et al., 2017), and have well-organized codes available. To achieve a comprehensive evaluation and analysis, we use nine satellite datasets from different locations, and each dataset contains ~100-500 stereo pairs with their respective ground truth LiDAR data. We train and test the models on the same and different datasets, and analyze the results to understand their performance, robustness, and generalization. To be more specific, this paper presents three contributions:

1. We comprehensively evaluate five stereo DIM algorithms (including four DL approaches) on satellite stereo images using hundreds of pairs from nine test-sites, to inform the community of the performance of such DIM algorithms under varying configurations.

2. We analyze the accuracy of the evaluated DL methods and study their robustness against stereo configurations of data that were reported to be critical for the resulting accuracy of DSMs for traditional methods (Qin, 2019).

3. We study the generalization capability (or transferability) of these DL stereo DIM methods trained on and applied to datasets across different geographical regions and resolution/sensors (including satellite, airborne and ground-view images).

The remainder of the paper is organized as follows: Section II introduces relevant work including a brief review of stereo DIM algorithms and existing comparative studies. Section III describes the dataset, preprocessing methods, and experimental data and setup as used in this work. Section IV describes in detail the dense stereo DIM algorithms and the evaluation method. Section V presents the results, evaluation, and discussion. Finally, the conclusion, limitations, and potential future directions are discussed in Section VI.

## II. RELATED WORKS

### A. Stereo DIM algorithms

There has been a tremendous development in stereo DIM algorithms over the years, they are broadly classified into traditional and DL methods (Zhou et al., 2020). Traditional methods are the very early algorithms with basic cost matching metrics as the sum of absolute differences (SAD), normalized cross-correlation (NCC), mutual information (MI), and Census transformation (Brown et al., 2003; Seitz et al., 2006). With the development of DL methods, the cost-matching pipeline was replaced by convolutional neural networks (CNN) (Žbontar & LeCun, 2015). DL-based algorithms have attracted great attention due to their superior performances in benchmark testing (Geiger et al., 2012). Depending on the task of learning, DL stereo methods can be further categorized into learning-based cost metrics (Žbontar & LeCun, 2015, 2016) and E2E learning (Chang & Chen, 2018; Cheng et al., 2020; Kendall et al., 2017; Xu & Zhang, 2020; Zhang, Prisacariu, et al., 2019). Learning-based cost metric was first introduced by (Žbontar & LeCun, 2015) to learn similarities from image patches. Both traditional and learning-based algorithms process low-level features as intensity or gradient patches to indicate similarity. As a result, their performance is limited to repetitive patterns and texture-less regions. Thus, post-processing like cost aggregation, optimization, and refinement based on these metrics is necessary to enhance the quality of the disparity map (Hirschmuller, 2005, 2008; Scharstein et al., 2001). DL methods rapidly evolved to E2E learning algorithms, where their main contribution is to replace the classical multistage optimization with a trainable network to directly predict the disparity from stereo images (Hamid et al., 2020; Laga, 2019). The underlying concept is that these neural networks can directly capture more global features, hence, they may better perform (Chang & Chen, 2018; Cheng et al., 2020; Kendall et al., 2017). In addition to these intensively studied methods, there are a few methods that perform context learning for part of the traditional pipeline but do not fully fall into either of these DL categories, for example, SGM-Net (Seki & Pollefeys, 2017) learns the per-pixel smoothness penalty and GA-Net (Zhang, Prisacariu, et al., 2019) learns networks to guide the cost-aggregation process.

*B. Existing comparative studies of stereo dense matching algorithms*

Most of the existing review papers on stereo DIM algorithms take upon single to few stereo pairs for evaluation (Hamid et al., 2020; Laga, 2019; Xia et al., 2020; Zhou et al., 2020), which may be insufficient to provide an accurate and conclusive evaluation. There are a few but limited studies concerning the use of more pairs to study DIM algorithms: they indicate that the quality of the DSM is significantly correlated with the radiometric and geometric characteristics and the configurations of the stereo pairs (Facciolo et al., 2017; Qin, 2019; Yan et al., 2016; X. Zhou & Boulanger, 2012). For instance, Facciolo et al., (2017) found that selecting stereo pairs based on specific heuristics as minimum seasonal differences improves DIM and reduces the uncertainties in the DSM. In (Qin, 2019), the author observed a direct relation between the geometric configurations at the time of acquisition as the sun angle difference and intersection angle (base-height ratio) and the accuracy of DSMs. In addition, the spatial distribution of objects in space (e.g., buildings density, sizes, and distances) and land cover types (e.g., trees, roads, water surfaces, etc.) are highly diverse across different test-sites (Chi et al., 2016), hence, may impact the performance of stereo DIM algorithms. With satellite images being rich in content and acquisition configurations, it is necessary to analyze the performance of stereo DIM algorithms on a large number of datasets covering a variety of regions, complexities, and configurations to understand their practical values in various applications.

*C. Deep learning training models and generalization*

Despite the superiority of DL algorithms in DIM, the generalization issue remains a major challenge. Learning across different domains such as data collected from different sensors, data of different locations, data with different spatial resolutions, etc., often leads to a deep drop in the accuracy of DIM because of their inability to predict disparity from unseen data (Pang et al., 2018; Song et al., 2021). One possible solution is to encapsulate a large number of training datasets covering all scenarios and instances that a network may encounter (Najafabadi et al., 2015). However, in practice obtaining a large training dataset with their ground truth (such as LiDAR) is costly and often unavailable for large-scale areas (Chi et al., 2016), it also requires manual or post-processing to convert to the ground-truth disparity which may produce systematic errors (Cournet et al., 2020; Patil et al., 2019; Song et al., 2021). Although some benchmark datasets are available (Bosch et al., 2016; Le Saux, 2019; Rottensteiner et al., 2012), existing evaluations are mostly performed on a dataset-by-dataset basis. Moreover, these datasets, although seem to be large in data volume, mostly present typical urban scenes of a few major cities and remain very sparse when considering the generalization in terms of scene contexts around the globe, sensor types, and resolutions.

### III. DATASET, PREPROCESSING, AND EXPERIMENTAL SETUP

In this section, we will first describe the datasets used in this work. Then, we will describe in detail the preprocessing methods and experimental setup applied to comprehensively train and evaluate the DL-based stereo matching algorithms.

*A. Dataset description*

In this work, we have two types of data that are used for training and evaluation purposes. All our data are collected from publicly available benchmarks. First, we will describe our evaluation datasets and then we will describe the training datasets.

The evaluation is based on satellite images from IARPA (Bosch et al., 2016) and the 2019 DFC (track 3) (Le Saux, 2019) benchmarks. They provide stereo images from the WorldView-3 satellite sensor with a spatial resolution of 0.3 meters. In addition, they provide the airborne LiDAR which we convert to the ground-truth DSM and use for evaluation. The IARPA benchmark provides 50 overlapping images for a 100 km$^2$ area near San Fernando, Argentina (ARG) collected between January 2015 and January 2016. The 2019 DFC benchmark provides 16 to 39 overlapping images for 100 km$^2$ of Omaha, NE, USA (OMA) and Jacksonville, FL, USA (JAX) collected from September 2014 to November 2015. We select three sub-regions as test-sites from OMA, JAX, and ARG datasets. Our selection covers test-sites with a variety of densities, complexities, land types, and covers. The 16-50 stereo images provided for all test-sites, yield approximately 6,278 stereo pair images. After omitting stereo pairs with extremely small intersection angles (i.e., < 2 degrees) and those fail during feature matching for relative orientation (Kuschk et al., 2014) due to large radiometric variations and clouds, we kept 2,861 stereo pairs for the experimental comparison. The stereo pair images have a wide range of geometric configurations. The sun angle difference ranges between 0 and 50 degrees and the intersection angle ranges between 0 and 67 degrees. Figure 1(a) provides detailed information about the evaluation dataset and test-sites.

The training dataset is used to train the DL algorithms. We include a variety of datasets from satellite, airborne, and ground-view sensors. For training with a satellite dataset, we use another set from the 2019 DFC benchmark known as track 2. The dataset provides 4,293 rectified stereo pair images of size 1024 × 1024 with their ground truth disparities derived from LiDAR. We also use datasets that are distinctively different from satellite images including ground-view images from *KITTI* dataset (Geiger et al., 2012) and airborne dataset from Toronto, Canada *ISPRS* (International Society for Photogrammetry and Remote Sensing) benchmark (Rottensteiner et al., 2012). The airborne dataset contains 13 images from an aerial block, captured using the UltraCam-D camera covering an area of 1.45 km$^2$ at a ground sampling distance (GSD) of 15 (cm). Each image is at a size of 11500 × 7500 pixels, and an intersection angle between neighboring images ranges from 15 to 30 degrees. The provided images have a varying overlap, thus in total, we have 8 stereo pair images. The **ISPRS** airborne dataset also provides LiDAR point clouds from Optech's airborne laser scanner (ALS), which we converted to ground-truth disparity and DSM for training and testing. The Toronto dataset includes a mixture of small to high-rise buildings, as well as other classes such as roads, trees, grass, etc. For more details refer to Figure 1(b).

| | | |
|---|---|---|
| **OMA I** 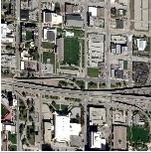 | **OMA II** 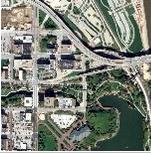 | **OMA III** 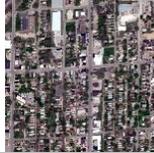 |
| Number of stereo pairs: 361<br>Image dimensions: 1168x1165<br>Sun angle difference range: 0°-27.7°<br>Intersection angle range: 1.53°-46.36°<br>Ground truth data: LiDAR | Number of stereo pairs: 293<br>Image dimensions: 1206x1202<br>Sun angle difference range: 0°-25.1°<br>Intersection angle range: 1.25°-45.89°<br>Ground truth data: LiDAR | Number of stereo pairs: 239<br>Image dimensions: 1187x1190<br>Sun angle difference range: 0°-25.3°<br>Intersection angle range: 1.43°-46.99°<br>Ground truth data: LiDAR |
| **JAX I** 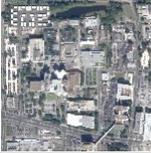 | **JAX II** 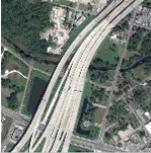 | **JAX III** 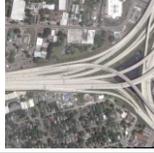 |
| Number of stereo pairs: 138<br>Image dimensions: 1163x1164<br>Sun angle difference range: 0.10°-50.90°<br>Intersection angle range: 2.64°-48.51°<br>Ground truth data: LiDAR | Number of stereo pairs: 91<br>Image dimensions: 1122x1152<br>Sun angle difference range: 0°-50.60°<br>Intersection angle range: 2.64°-48.51°<br>Ground truth data: LiDAR | Number of stereo pairs: 247<br>Image dimensions: 1133x1164<br>Sun angle difference range: 0°-55.30°<br>Intersection angle range: 2.64°-53.22°<br>Ground truth data: LiDAR |
| **ARG I** 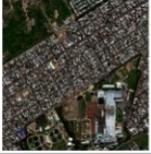 | **ARG II** 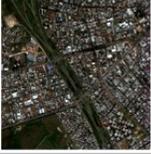 | **ARG III** 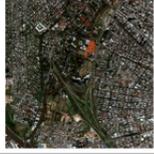 |
| Number of stereo pairs: 503<br>Image dimensions: 2565x2561<br>Sun angle difference range: 0.04°-49.02°<br>Intersection angle range: 1.42°-66.20°<br>Ground truth data: LiDAR | Number of stereo pairs: 475<br>Image dimensions: 4199x4256<br>Sun angle difference range: 0.02°-49.04°<br>Intersection angle range: 3.26°-54.55°<br>Ground truth data: LiDAR | Number of stereo pairs: 472<br>Image dimensions: 4062x4208<br>Sun angle difference range: 0.02°-49.04°<br>Intersection angle range: 1.44°-60.90°<br>Ground truth data: LiDAR |

(a) Satellite datasets stereo pair images information for OMA, JAX, and ARG test-sites.

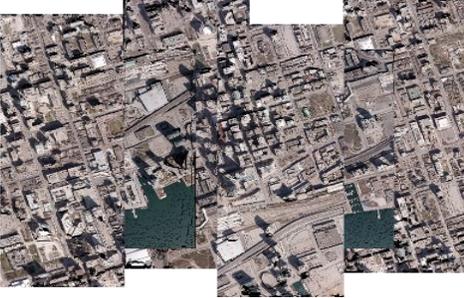

Number of stereo pairs: 8
Image dimensions: 11500 x 7500
Intersection angle range: 15°-30°
Ground truth data: LiDAR

(b) Airborne dataset stereo pair images information for Toronto, Canada test-site.

Figure 1. Information about the stereo pair images in the (a) satellite and (b) airborne datasets.

*B. Ground truth data derivation and preprocessing*

The ground-truth disparities for the satellite and airborne datasets are derived from LiDAR point clouds following three steps (Patil et al., 2019), first, the point clouds are aligned to the original unrectified images using correlating methods to register the images and the LiDAR intensity data, such as using the mutual information (MI). The offsets between these two sources are further used to adjust their corresponding RPC parameters of the satellite images; second, the stereo pair images are rectified using their adjusted RPC models to the epipolar image space; third, the 3D LiDAR point clouds are projected to the satellite images and mapped to the rectified left and right images; finally, the disparity map is computed for both the left and right rectified image based on the projected 3D points. Obtaining disparity maps with sub-pixel accuracy from the LiDAR on the satellite images is very challenging. According to Wu et al. (2021), the transformation between LiDAR and ground truth disparity may produce additional errors. Whereas, Patil et al. (2019) report that the ground truth disparity in the 2019 DFC benchmark (Bosch et al., 2019) lacks adequate validation, thus may include errors that have not been acknowledged. A common source of error is the change of scene content between LiDAR and satellite images due to different acquisition times. As can be noticed in the example in Figure 2, the yellow rectangles show some trees in the ground truth disparity that are not recorded in the image. Other sources of errors include missing points in the ground truth disparity due to occlusions or shadows (see a blue rectangle in Figure 2), random noise and artifacts (see red and green rectangles in Figure 2), or systematic errors from the image rectification process. These errors will inevitably impact on the training of the DL models. Therefore, we minimize some of these errors by pre-filtering the disparity map using a 3-sigma rule, and we fill the small holes according to the nearest neighbors to create a smooth disparity map for training purposes.

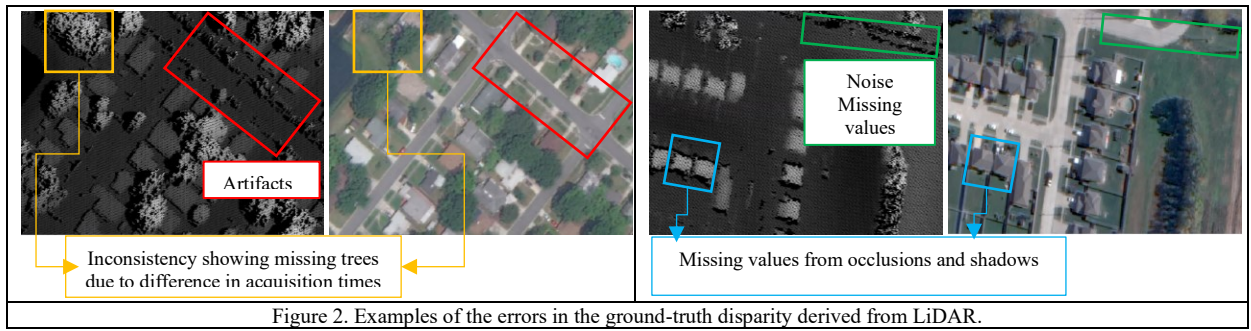

Figure 2. Examples of the errors in the ground-truth disparity derived from LiDAR.

*C. Experimental data and setup*

1) *Training data*

The satellite training dataset is from the 2019 DFC (track 2) benchmark. It has 4,293 rectified stereo images of size 1024 × 1024 pixels with their ground truth disparity maps. Because of limited GPU memory, we crop the images to small patches with W=1248 pixels and H=384 pixels, yielding a total number of 25,705 training patches. We then split the input data into 80% training and 20% testing. The airborne training dataset includes Toronto dataset from the ISPRS benchmark. There are 8 stereo images with sufficient overlap, each with a size of 11500 × 7500 pixels. We rectify these images and crop them to small patches of size W=1248 pixels and H=384 pixels. The total number of patches is around 3000.
There are two types of DL methods based on our taxonomy (i.e., learning-based and E2E learning), and both are trained using two different datasets. MC-CNN is supposed to be trained using both the satellite and the airborne dataset. However, considering that the ground-level KITTI dataset should be more distinctively different, and it has been well-trained already as shared in its original work (Žbontar & LeCun, 2016), we therefore alternatively use the KITTI dataset and satellite dataset as more challenging testing. The same was attempted for E2E, while this overwhelmed the E2E algorithms to yield any meaningful results. Therefore, we only train the E2E networks using satellite and airborne datasets.

2) *Training setup*

The training parameters and settings for each DL network are shown in Table I. The settings include preprocessing steps such as color normalization and random crop of the input images to smaller patch size, in addition to the parameters selected prior to training such as the type of optimizer, learning rate, batch sizes, etc.

Table I. The training parameters and setup for the DL models

| Input parameters | MC-CNN | GCNet[*] | PSMNet[*] | LEAStereo[*] |
|---|---|---|---|---|
| Color normalization | Grayscale patches | RGB normalization | RGB normalization | RGB normalization |
| Patch size | 9 × 9 | 512 × 256 | 512 × 256 | 192 × 384 |
| Optimizer | Gradient Descent | Adam | Adam | Stochastic Gradient Descent |
| Optimization parameters | β = 0.9 | β1 = 0.9, β2 =0.999 | β1 = 0.9, β2 = 0.999 | β = 0.9 |
| Learning rate (λ) | λ=0.002 decay: 0.9 (after 10 epochs) | λ=0.001 | λ=0.001 | Cosine λ= 0.025 - 0.001 decay: 0.0003 |
| Maximum disparity | - | 192 | 192 | 192 |
| Batch size | 1 | 1 | 1 | 1 |
| # Epochs for training using satellite dataset | 10 | 35 | 10 | 57 |
| # Epochs for training using airborne dataset | - | 500 | 20 | 300 |

[*]*E2E learning algorithms are implemented in Pytorch and trained on Nvidia GeForce RTX 2080 Ti.*

## IV. THE DENSE STEREO MATCHING METHODS FOR EVALUATION

In this section, we describe the main stereo matching approaches used in our analysis. First, we will briefly present the selected stereo DIM algorithms including: 1) Census+SGM, 2) MC-CNN+SGM, and 3) E2E methods using GCNet, PSMNet, and LEAStereo. Then, we will describe the evaluation method as applied in our analysis. In particular, MC-CNN has two architectures called MC-CNN-fst and MC-CNN-acc (Žbontar & LeCun, 2016); the former uses a product of two feature vectors to decide the similarity and the latter uses fully connected layers following the concatenated feature vectors. Although the latter was reported to be slightly more accurate, we use the former version (MC-CNN-fst) in our evaluation, given its good leverage of performance and spend for large-sized satellite images. Without specifically stated, MC-CNN mentioned hereafter refers to the MC-CNN-fst version.

### A. Stereo DIM algorithms

As mentioned in Section II-A, We evaluate two types of DL approaches, 1) DIM methods following a classic multi-stage paradigm including feature extraction, cost matching, and cost aggregation for disparity computation, where DL component can be one of these stages, for example, the learning-based cost metric in our evaluation learns the feature extraction, or 2) DL methods with E2E learning which use trainable networks to learn from stereo data and ground-truth disparities, and directly predict disparity maps. All approaches require rectified epipolar images as inputs, which we generate using RPC Stereo Processor (RSP) software (Qin, 2016). As mentioned in Section I, we use SGM for cost aggregation following Census and MC-CNN cost metrics as proposed by (Hirschmuller, 2005, 2008). In this subsection, we briefly introduce these stereo DIM algorithms used for our evaluations.

1) *Cost metrics for classic multi-stage stereo DIM methods*

***Census*** maps a window of pixels across left and right images to find similar features with minimum cost (Hirschmuller, 2005; Zabih & Woodfill, 1994). First, a fixed window (*w*) is set around every pixel in the left image and a sliding window scans the same epipolar line in the right image. The pixels in *w* are transformed into a vector of binary strings. Depending on their values relative to the central pixel, a value of 1 is assigned if a pixel is larger than the central pixel and 0 otherwise. The binary vectors are processed by a hamming distance and summed to compute the cost and indicate the number of variant pixels. A match is selected based on the minimum cost. A detailed example of the steps is shown in Figure 3(a).

***MC-CNN*** uses a Siamese convolutional neural network to compute the matching cost as shown in Figure 3(b). The training is performed using small patches of size 9×9 extracted from grayscale left and right images and their ground truth disparity maps. The testing, on the other hand, takes the entire image as input. During training, the network learns from each pair of patches by classifying them into similar or dissimilar. The patches are processed separately in two sub-networks that include four convolutional layers with a rectified linear activation function (ReLU), a convolutional layer, and a normalization layer. The sub-networks generate two normalized feature vectors that are fed into a cosine dot product to compute their similarity and indicate a good or bad match. In terms of model parameters, MC-CNN has a small model capacity with a total number of parameters of 0.3 million (M).

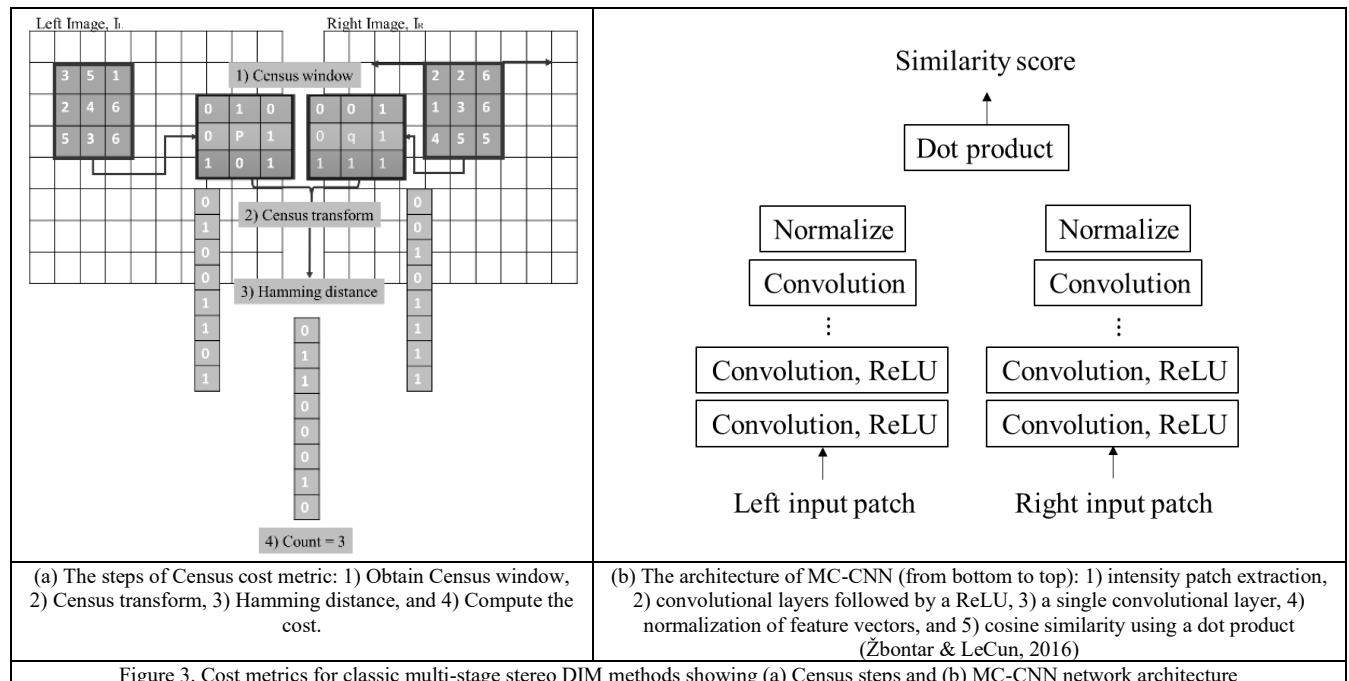

| (a) The steps of Census cost metric: 1) Obtain Census window, 2) Census transform, 3) Hamming distance, and 4) Compute the cost. | (b) The architecture of MC-CNN (from bottom to top): 1) intensity patch extraction, 2) convolutional layers followed by a ReLU, 3) a single convolutional layer, 4) normalization of feature vectors, and 5) cosine similarity using a dot product (Žbontar & LeCun, 2016) |

Figure 3. Cost metrics for classic multi-stage stereo DIM methods showing (a) Census steps and (b) MC-CNN network architecture

2) *E2E learning stereo DIM methods*

Three recently published E2E networks (with available codes) are described in this subsection. These methods vary in their model architecture, backbone building blocks, and model complexity (model capacity). Some of the architectures are designed to readily model the conceptual nature of depth sensing and recognition, such as the pyramid architectures (Chang & Chen, 2018) that process the data in a coarse-to-fine fashion. Understanding the model capacity is useful to determine the degree of generalization and applicability in practice. Generally, a network with a large number of parameters reduces the training error, increases the computational cost, and most likely has a high testing error, thus, leading to overfitting and poor generalization problems (Huang et al., 2021).

***GCNet*** involves several steps to generate the disparity map. It begins with unary feature extraction from the left and right images using 2D convolutional layers (CNNs) and eight residual blocks. Then, it matches these features at every disparity level by concatenating both feature maps to build the 4D cost volume. The cost volume is then regularized via cost aggregation using 3D CNNs to learn contextual information and refine the cost volume. Finally, it applies regression to predict the final disparity map from the regularized cost volume (Kendall et al., 2017). The total number of model parameters for GCNet is 3.5M, which is the highest among the three selected E2E algorithms. For more details on the network's architecture see Figure 4.

***PSMNet*** starts by extracting feature maps using three CNN layers and four residual blocks. These features are then processed hierarchically on multiple scales through Spatial Pyramid Pooling (SPP) module to generate the 4D cost volume. The SPP module can enhance the visibility and feature matching of deformed objects (He et al., 2014). The cost volume is then processed by a stacked 3D hourglass module for cost aggregation and regularization, which can provide better learning of the context information. The final step is to apply regression to estimate the disparity map (Chang & Chen, 2018). PSMNet has a total number of model parameters of 2.8M. For more details on the network's architecture see Figure 4(b).

***LEAStereo*** uses a hierarchical neural architecture search (NAS) pipeline to find the optimal stereo matching network parameters (Cheng et al., 2020). The optimal parameters determined by NAS may include the filter size of the convolutional layer, strides, etc. The stereo matching network consists of two sub-networks, first, the feature net, which is used to extract the features from the left and right image to generate the 4D feature volume, which is then processed by the matching net to compute the matching cost (using 3D convolutions) to generate the 3D cost volume. Finally, regression is applied to compute the final estimates of the disparity map. LEAStereo has the least number of model parameters about 1.81M. For more details on the network's architecture see Figure 4.

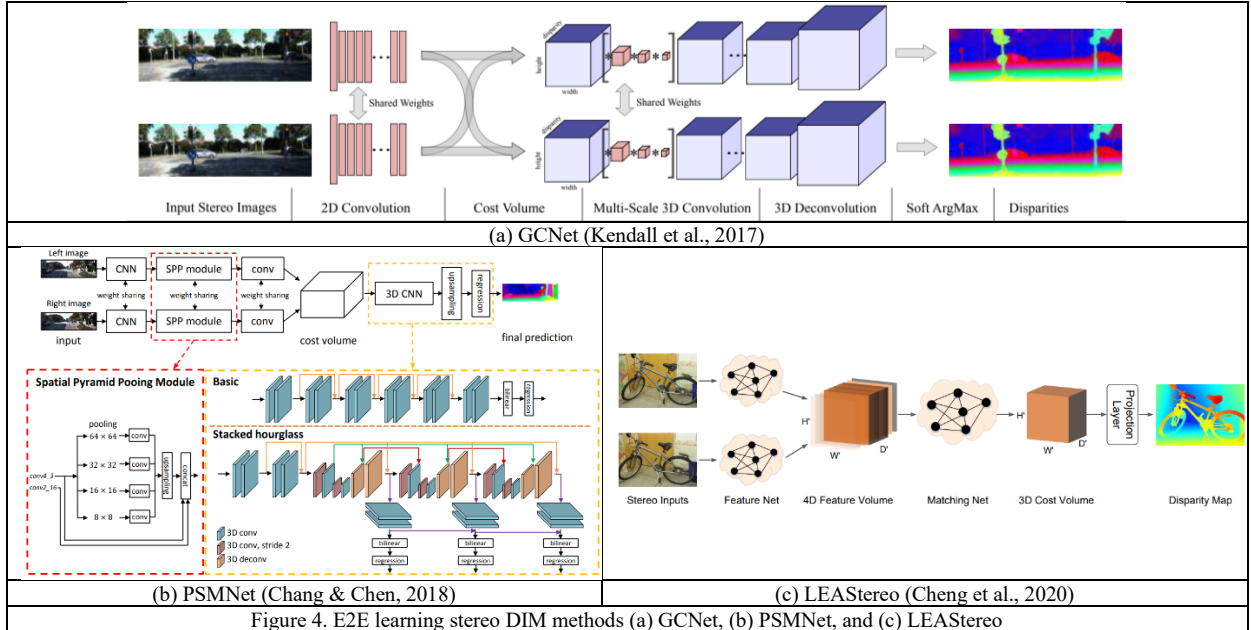

Figure 4. E2E learning stereo DIM methods (a) GCNet, (b) PSMNet, and (c) LEAStereo

*B. Evaluation of the stereo DIM*

For evaluation and comparison of the stereo DIM algorithms, we triangulate the predicted disparity to DSMs. We assess each DSM *I(x,y)* against the ground-truth DSM from LiDAR *GT(x,y)* using the root mean squared error (RMSE) as follows:

$$RMSE = \sqrt{\frac{\sum_{i,j=1}^{N,M}(I(x,y) - GT(x,y))^2}{N \times M}} \quad (1)$$

Where *(x,y)* are the pixel positions and *(N, M)* are the DSM dimensions.

## V. EXPERIMENTS RESULTS AND DISCUSSION

Based on the DSMs generated using different stereo DIM algorithms for the entire nine test-sites, we assess each class of the algorithms in the following three aspects:

1) Overall performance: we evaluate the RMSE of the resulting DSMs of each algorithm on each of the few hundred pairs based on the average RMSE, error distribution, completeness rate, and visual analysis.

2) The robustness with respect to varying acquisition configurations: we assess for each algorithm, the impact of two critical geometric/radiometric parameters, i.e., sun angle difference and intersection angle, on the resulting DSM through correlation analysis using hundreds of stereo pairs.

3) Generalization capability of DL algorithms: we evaluate for each of the DL methods, their RMSE against the ground truth by training the model on one dataset and applying it to another dataset (of a different location and/or different sensor).

Note that co-registration can be a critical factor for this evaluation as well as many other applications such as change detection and multimodal data fusion (Lv et al., 2021, 2022). To achieve reasonable co-registration, we first aligned the DSMs to a reference DSM. Following the method proposed by Gruen and Akca (2005), we applied a simple version of least-squares matching to optimize the shifts in all three dimensions. Blunders are eliminated progressively by using a predefined threshold of 5 meters, computed as a factor of 10 times of the ground sampling distance. DSMs are considered well-registered and unbiased when their mean difference to the reference DSM is smaller than a pixel-size, i.e., 0.5 meters.

*A. The overall performance analysis*

Performance analysis is the key factor to assess the reliability of stereo DIM algorithms. We use standard statistical measures such as the average RMSE and distribution of errors to analyze the DSMs from the 2,681 stereo pairs over the nine test-sites. We are particularly interested in the achievable upper and lower bounds of the accuracy for all stereo pairs. We also provide an analysis of the completeness rate and visual results. Based on knowledge from prior literature, we expect DL algorithms to outperform traditional methods (i.e., Census+SGM) since they are more complex and trained on examples of ground-truth data.

Table II(a) presents the average errors (RMSE) of the DSMs from all stereo pairs and test-sites. We can notice that E2E algorithms can achieve the minimum errors. This can be obvious from several aspects. First, they have minimum overall average errors as shown from PSMNet and LEAStereo where their values are 4.73 and 4.83 meters, respectively (see the last row in Table II(a)). Second, eight of nine test-sites have the lowest errors in one of GCNet, PSMNet, and LEAStereo (indicated in bold in Table II(a)). LEAStereo in particular has the most frequent lowest average RMSE among all testing sites. We can also notice that E2E algorithms can achieve the highest errors in some test-sites. Their performance varies drastically across different test-sites. Their average errors can range from 2 to 18 meters. For instance, the average RMSE for GCNet, PSMNet, and LEAStereo in JAX II is 2.73, 2.76, and 4.39 meters, respectively, while for OMA II it is high as 17.56, 8.99, and 8.84 meters, respectively. This implies that E2E methods do not predict disparity maps well from unseen data. Census+SGM and MC-CNN+SGM show opposite performance to E2E methods. They have more consistent average errors across different test-sites close to the overall average errors (See Table II(a)). In addition, Census+SGM outperforms MC-CNN+SGM, as it achieves a lower overall average RMSE of 5.50 meters, as compared to 6.43 m achieved by MC-CNN+SGM (see Table II(a)). This result reveals a different conclusion from the original MC-CNN (Žbontar & LeCun, 2015), which may be due to the sensitivity of DL algorithms towards noise in the ground-truth disparity maps used for training the networks (as indicated in Section III-B).

While E2E methods can achieve the lower and upper bounds of the average errors, their average standard deviation in Table II(b) indicates higher consistency for the error distribution of stereo pairs in the same test-site. GCNet, PSMNet, LEAStereo have the lowest average standard deviations (less than 2.73 meters) as can be seen in the last row in Table II(b). This is because of their ability to learn similar context and geometry, which makes them invariant to changing radiometric properties of the stereo pairs. In contrast, Cenus+SGM and MC-CNN+SGM have higher standard deviations with averages of 4.08 and 3.40 meters, respectively. This implies that similarity-learning algorithms are sensitive to the radiometric properties of the stereo pair images.

In general, E2E methods are able to achieve the absolute minimum errors. Their ability to learn from context makes them robust to varying radiometric properties of the stereo pairs. However, they can also have the absolute highest errors when learning from new unseen test-sites. In contrast, traditional and learning-based algorithms may have higher errors but can provide more consistent performance across different test-sites.

Table II. The overall performance of the stereo DIM algorithms over nine test-sites presented by (a) average RMSE, (b) standard deviation of the RMSE, and (c) completeness rates of the DSMs. Note: The bold numbers in (a) and (b) indicate the lowest values of the average RMSE and standard deviations, and in (c) they indicate the highest completeness rates.

| Test-site ID | (a) Average RMSE (meters) | | | | | (b) Standard deviation of the RMSE (meters) | | | | |
|---|---|---|---|---|---|---|---|---|---|---|
| | Census +SGM | MC-CNN +SGM | GCNet | PSMNet | LEAStereo | Census +SGM | MC-CNN +SGM | GCNet | PSMNet | LEAStereo |
| *OMA I* | 6.55 | 5.94 | 18.11 | **4.35** | 5.20 | 6.32 | 4.16 | 9.63 | **3.19** | 3.49 |
| *OMA II* | **6.52** | 7.46 | 17.56 | 8.99 | 8.84 | 4.71 | 4.31 | 7.61 | 5.22 | 4.77 |
| *OMA III* | 3.32 | 3.96 | 3.30 | 2.55 | **2.14** | 2.20 | 2.04 | 1.47 | 0.75 | **0.50** |
| *JAX I* | 5.75 | 10.60 | 6.79 | **5.46** | 5.97 | 1.35 | 4.52 | 2.31 | 1.48 | 1.47 |
| *JAX II* | 5.19 | 7.32 | 3.31 | 3.30 | **3.23** | 4.72 | 3.03 | **0.19** | 0.21 | **0.19** |
| *JAX III* | 4.29 | 7.07 | **2.73** | 2.76 | 4.39 | 3.92 | 5.08 | **0.31** | 0.33 | 0.46 |
| *ARG I* | 5.31 | 5.07 | 4.25 | 5.03 | **3.88** | 4.14 | 3.39 | **1.16** | 4.08 | 2.21 |
| *ARG II* | 5.84 | 5.01 | 5.13 | 5.00 | **4.84** | 4.63 | 1.27 | **0.80** | 3.99 | 3.44 |
| *ARG III* | 6.69 | 5.42 | **4.78** | 5.08 | 4.98 | 4.72 | 2.76 | **1.00** | 1.73 | 1.94 |
| *Overall Average error* | 5.50 | 6.43 | 7.33 | 4.73 | **4.83** | 4.08 | 3.40 | 2.72 | 2.33 | **2.05** |

| Test-site ID | (c) Average completeness (%) | | | | |
|---|---|---|---|---|---|
| | *Census +SGM* | *MC-CNN +SGM* | *GC-Net* | *PSM-Net* | *LEA-Stereo* |
| OMA I | 89.29 | 87.94 | 92.53 | 92.42 | **97.42** |
| OMA II | 73.31 | 70.71 | 87.57 | 87.45 | **95.41** |
| OMAIII | 74.13 | 86.94 | 90.11 | 87.28 | **90.33** |
| JAX I | 90.29 | 86.37 | 93.6 | **93.54** | 93.54 |
| JAX II | 86.22 | 90.99 | 94.4 | **94.47** | 94.38 |
| JAX III | 87.08 | 81.71 | 95.64 | 95.28 | **99.29** |
| ARG I | 75.13 | 74.84 | 94.78 | 93.79 | **94.86** |
| ARG II | 63.1 | 79.22 | 89.99 | 89.23 | **95.74** |
| ARGIII | 81.53 | 87.66 | **94.32** | 87.08 | 93.81 |
| *Average* | 80.01 | 82.93 | 92.55 | 91.17 | **94.98** |

The completeness rate is another assessment metric for performance. It refers to the number of filled pixels over the entire testing site. It indicates two main components: first, the degree to which an algorithm is able to infer disparity values in all situations like the existence of insufficient texture, occlusions, radiometric inconsistencies, etc.; second, the percentage of correctly matched points relative to the number of filled pixels. The completeness rates are shown in Table II(c). We can notice that E2E learning algorithms have the highest completeness rates (>91%), with LEAStereo in specific having the absolute highest completeness rates with an average of around 95%. On the other hand, traditional and learning-based methods have lower completeness rates in the lower 80% range. The visual illustration in Figure 5 presents sample DSMs from ARG I test-site, where the white pixels represent missing values. We can see that the DSMs from Census+SGM and MC-CNN+SGM have more missing values and less distinctive objects. Whereas, the DSMs from E2E algorithms have more filled pixels and a higher level of details with better visibility of objects as indicated by the boundaries of the building.

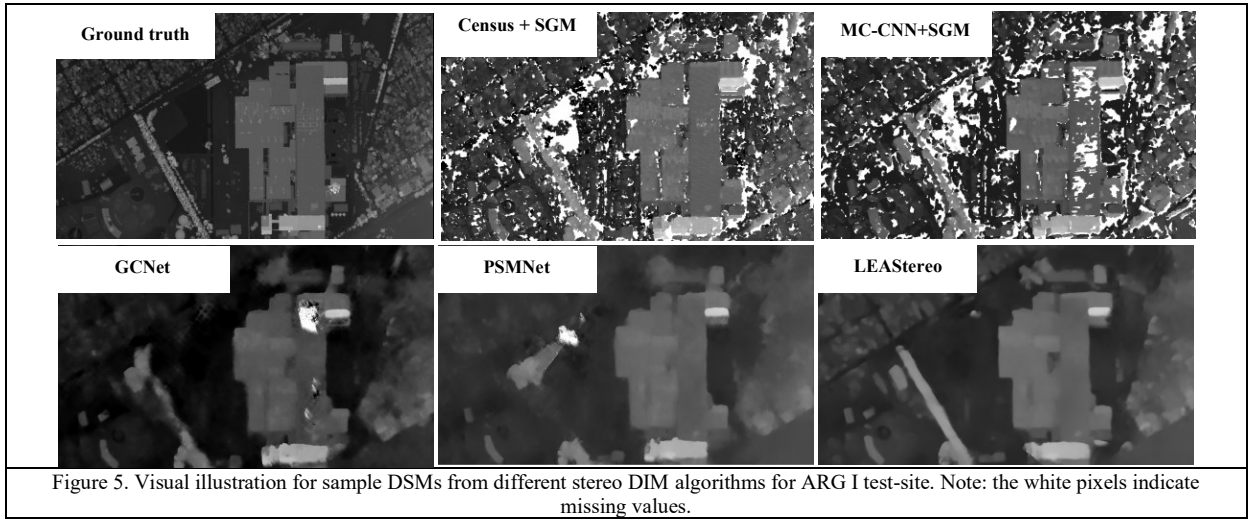

Figure 5. Visual illustration for sample DSMs from different stereo DIM algorithms for ARG I test-site. Note: the white pixels indicate missing values.

### B. Robustness analysis using varying geometric configurations of stereo images pairs

The robustness measures the ability of the stereo DIM algorithm to maintain stable performance across stereo pairs with varying acquisition configurations. As reported by (Qin, 2019), geometric configuration parameters like the sun angle difference and intersection angle have a critical impact on the accuracy of DSMs generated using Census+SGM. Similarly, we aim to study their impact on the results of the DL stereo DIM algorithms. We use robust regression to fit a model between the RMSE of the DSMs and the geometric parameters (for all stereo pairs) to measure the correlation and eliminate potential outliers. The value of $R^2$ ($\in [0,1]$) represents the goodness of fit and the degree of impact. Thus, we use $R^2$ to indicate the amount of impact the geometric parameters have on the accuracy of DSMs as follows:

$$R^2 = \frac{S_0^2 - S_1^2}{S_0^2} \quad (2)$$

With $S_0^2 = \sum_{i=1}^{n}(y_i - \bar{y})^2$ and $S_1^2 = \sum_{i=1}^{n} r_i^2$, where $S_0^2$ is the sum of squares between the input observations $y_i$ and the mean $\bar{y}$, and $S_1^2$ is the sum of residuals $r_i$ computed from the difference between the RMSE fitted and predicted from the regression model and the actual computed RMSE. Given previous observations in (Qin, 2019), we expect DL algorithms to be impacted by varying acquisition configurations.

Table III presents $R^2$ showing the relationship between the RMSE of the DSMs and the geometric parameters for all stereo pair images in the test-sites. Census+SGM shows the highest $R^2$ scores between the RMSE of DSMs and each of the sun angle difference and the intersection angle with an average of 0.52 and 0.53, respectively. In addition, five to six test-sites have $R^2$ values between 0.53 and 0.82. This implies that traditional DIM methods are impacted by the geometric parameters. As for all DL algorithms, $R^2$ scores with respect to the sun angle difference are very small and their average does not exceed 0.37. While for $R^2$ with respect to the intersection angle, the learning-based MC-CNN is the least impacted by the intersection angle where it has the minimum average $R^2$ score of 0.36. On the other hand, the $R^2$ for E2E algorithms slightly increase to an average of 0.38, 0.45, and 0.48 for GCNet, PSMNet, and LEAStereo, respectively. Despite this minor increase, it is still insignificant in comparison to Census+SGM. This indicates that DL stereo DIM algorithms can be more robust to varying geometric configuration parameters that are associated with the stereo pair images.

Table III. Robust regression analysis denoted by $R^2$ to indicate the relationship between the DSM accuracy and geometric configuration parameters.

| Test-site ID | $R^2$ with respect to the sun angle difference | | | | | $R^2$ with respect to the Intersection angle | | | | |
|---|---|---|---|---|---|---|---|---|---|---|
| | Census +SGM | MC-CNN +SGM | GCNet | PSMNet | LEAStereo | Census +SGM | MC-CNN +SGM | GCNet | PSMNet | LEAStereo |
| OMA I | **0.80** | 0.50 | 0.02 | 0.53 | 0.39 | **0.82** | 0.50 | 0.23 | 0.55 | 0.44 |
| OMA II | **0.27** | 0.26 | 0.03 | 0.13 | 0.08 | 0.30 | 0.27 | **0.53** | 0.25 | 0.28 |
| OMA III | **0.31** | 0.20 | 0.13 | 0.17 | 0.11 | **0.53** | 0.29 | 0.40 | 0.10 | 0.06 |
| JAX I | **0.35** | 0.26 | 0.01 | 0.00 | 0.00 | 0.33 | 0.18 | **0.74** | 0.72 | 0.73 |
| JAX II | **0.53** | 0.03 | 0.21 | 0.08 | 0.09 | **0.48** | 0.12 | 0.21 | 0.19 | 0.43 |
| JAX III | **0.63** | 0.41 | 0.07 | 0.07 | 0.03 | **0.66** | 0.42 | 0.11 | 0.27 | 0.32 |
| ARG I | 0.62 | **0.67** | 0.11 | 0.50 | 0.58 | 0.60 | 0.57 | 0.14 | 0.49 | **0.61** |
| ARG II | 0.41 | 0.23 | 0.52 | **0.68** | 0.67 | 0.42 | 0.17 | 0.43 | **0.67** | 0.66 |
| ARG III | 0.76 | 0.71 | 0.70 | **0.80** | 0.79 | 0.62 | 0.69 | 0.67 | **0.81** | 0.77 |
| Average | **0.52** | 0.37 | 0.20 | 0.33 | 0.30 | **0.53** | 0.36 | 0.38 | 0.45 | 0.48 |

Note: Bold indicates the highest $R^2$ values. All DL models are trained using the satellite dataset from the 2019 DFC benchmark.

C. The model generalization analysis for the DL stereo DIM algorithms

DL algorithms have difficulties generalizing to new unseen scenes. This is because of the significant differences in the characteristics of images such as intensities, illuminations, scene content, textures, noise level, etc. DL algorithms may have different generalization capabilities. To evaluate their generalization, we use similar and different training and testing datasets i.e., from different sensors and regions.

As mentioned in Section III-C, due to the different architectures of the DL algorithms, we train and test the learning-based and E2E methods on similar and different datasets. Figure 6. illustrates the training and testing pipelines. For simplicity, we are going to refer to the 2019 DFC (track 2) dataset as the "satellite-training" dataset and for the OMA, JAX, and ARG datasets as the "satellite-testing" dataset. We train MC-CNN using KITTI and satellite-training datasets separately and test it on the satellite-testing dataset from the nine test-sites. While for E2E algorithms, we train them separately on airborne and satellite-training datasets and test them on the airborne dataset. Given that MC-CNN is a binary cost metric, we expect it will have good generalization. As for E2E algorithms, since we use similar training and testing datasets (i.e., similar scene content as roads, buildings, etc.), we expect them to generalize well.

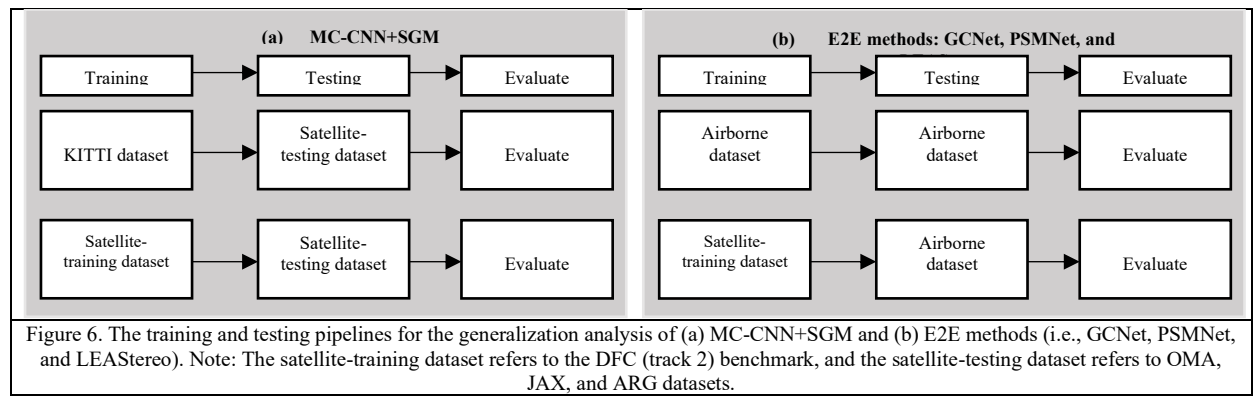

Figure 6. The training and testing pipelines for the generalization analysis of (a) MC-CNN+SGM and (b) E2E methods (i.e., GCNet, PSMNet, and LEAStereo). Note: The satellite-training dataset refers to the DFC (track 2) benchmark, and the satellite-testing dataset refers to OMA, JAX, and ARG datasets.

In Table IV(a), we compare the DSMs from MC-CNN+SGM trained independently on KITTI and satellite-training datasets and applied to the satellite-testing datasets from the nine test-sites. We observe a slight difference between the overall average RMSE of the two training models of 0.29 meters, with MC-CNN+SGM trained using the KITTI dataset performing slightly better. In addition, five test-sites have a lower average RMSE for the model trained using KITTI dataset than the satellite-training dataset. This may be due to some imperfections in the ground-truth disparity derived from LiDAR as indicated in Section III-B. Census+SGM seems to outperform MC-CNN+SGM (trained using KITTI and satellite-training dataset) in terms of the number of test-sites with the minimum average RMSE. This implies that Census-SGM is more stable than MC-CNN across different test-sites. Figure 7 shows an example for DSMs from both training models for OMA I test-sites. It is obvious there is a large resemblance in the DSMs, however in OMA I, we can notice some errors randomly scattered over the DSM generated from the KITTI trained model. The results show a good generalization of MC-CNN+SGM and that it can perform well on different datasets. Furthermore, it indicates the importance of the quality of the training data which can directly impact the results and must be given greater attention.

In Table IV(b), we compare DSMs from E2E algorithms trained on each airborne and satellite-training dataset separately and applied to the airborne dataset. It is obvious that there is a significant jump in the average RMSE for GCNet and LEAStereo of about 20 to 45 meters difference between the results of airborne and satellite training models, where PSMNet shows about 6 meters difference. This indicates that the generalization degree varies based on the model architectures, and PSMNet shows significantly better results. Part of the reason is that its pyramid architecture can better adapt data with different resolutions and scales than the other two. Additionally, E2E algorithms trained on and applied to airborne datasets outperform traditional and learning-based algorithms, as they can achieve the lowest average errors (RMSE) and the most frequent minimum errors. The results from Table IV(b) indicate that the source of the training data plays an important role. Example results are shown in Figure 6 for GCNet, PSMNet, and LEAStereo trained using airborne and satellite-training datasets and applied to the airborne dataset. It shows LEAStereo and GCNet perform poorly when trained using the satellite-training datasets. The PSMNet reaches better results but presents many noises and outliers (see red circles in Figure 8). This implies that E2E algorithms have a poor generalization capability and require domain-specific training datasets to achieve acceptable performance.

Table IV. Generalization analysis of (a) MC-CNN+SGM and (b) E2E algorithms.

(a) MC-CNN+SGM trained using satellite-training or KITTI datasets and tested on satellite datasets.

| Test-site ID | (a) Average RMSE (meters) | | | (b) Standard deviation for the RMSE (meters) | | |
|---|---|---|---|---|---|---|
| | Census +SGM | MC-CNN + SGM | | Census +SGM | MC-CNN + SGM | |
| | | Trained using satellite-training dataset | Trained using KITTI dataset | | Trained using satellite-training dataset | Trained using KITTI dataset |
| OMA I    | 6.55 | **5.94** | 7.55  | 6.32 | **4.16** | 6.77 |
| OMA II   | **6.52** | 7.46 | 7.28 | 4.71 | **4.31** | 5.02 |
| OMAIII   | **3.32** | 3.96 | 3.81 | 2.20 | **2.04** | 2.53 |
| JAX I    | **5.75** | 10.60 | 6.44 | **1.35** | 4.52 | 1.70 |
| JAX II   | **5.19** | 7.32 | 5.38 | 4.72 | **3.03** | 3.68 |
| JAX III  | **4.29** | 7.07 | 5.41 | **3.92** | 5.08 | 4.26 |
| ARG I    | 5.31 | **5.07** | 6.84 | 4.14 | **3.39** | 5.90 |
| ARG II   | 5.84 | **5.01** | 6.32 | 4.63 | **1.27** | 3.57 |
| ARGIII   | 6.69 | **5.42** | 6.21 | 4.72 | **2.76** | 3.96 |
| Overall Average | 5.50 | 6.43 | **6.14** | 4.08 | **3.40** | 4.15 |

*Note: Bold font indicates the minimum values.*
*The satellite-training dataset refers to the 2019 DFC satellite benchmark.*

(b) E2E algorithms trained using satellite-training or airborne datasets and tested on airborne datasets.

| Stereo pair ID | Census +SGM | MC-CNN+SGM | Trained using airborne dataset | | | Trained using satellite-training dataset | | |
|---|---|---|---|---|---|---|---|---|
| | | | GCNet | PSMNet | LEAStereo | GCNet | PSMNet | LEAStereo |
| 1 | **15.34** | 18.84 | 15.89 | 17.39 | 15.63 | 34.52 | 19.01 | 65.72 |
| 2 | 14.99 | 17.82 | 14.66 | 14.86 | **14.42** | 35.20 | 18.00 | 67.65 |
| 3 | **8.13** | 11.44 | 9.08 | 9.87 | 11.49 | 36.48 | 12.04 | 74.71 |
| 4 | 11.37 | **11.32** | 12.40 | 12.40 | 19.72 | 28.47 | 19.72 | 56.49 |
| 5 | 21.49 | 29.00 | 12.96 | 16.52 | **12.50** | 29.52 | 13.85 | 56.21 |
| 6 | **6.17** | 7.78 | 6.51 | 6.98 | 7.65 | 34.52 | 41.24 | 73.56 |
| 7 | 30.41 | 29.58 | 24.32 | 28.42 | **23.34** | 29.46 | 24.62 | 29.46 |
| 8 | 23.98 | 22.62 | 20.35 | 20.72 | **20.15** | 55.41 | 22.02 | 55.41 |
| Average RMSE (meters) | 16.49 | 18.55 | **14.52** | 15.90 | 15.61 | 35.45 | 21.31 | 59.90 |

*Note: Bold indicates the minimum values.*
*MC-CNN+SGM is trained using the 2019 DFC satellite benchmark.*
*The satellite-training and airborne datasets used to train E2E models refer to the 2019 DFC and Toronto ISPRS benchmarks.*

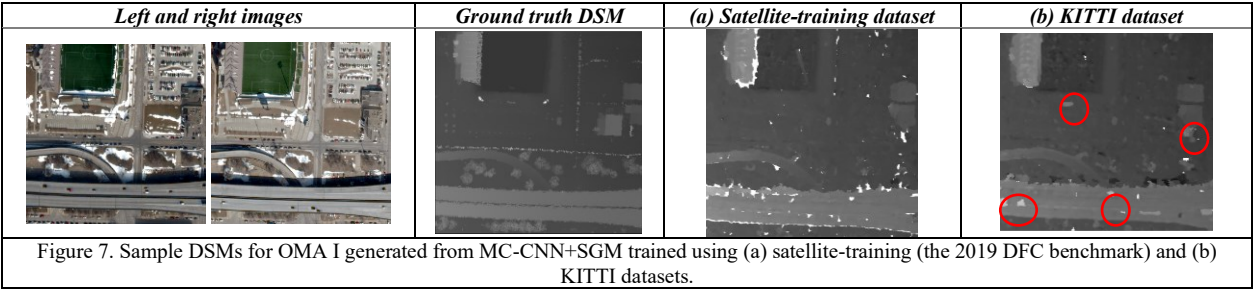

Figure 7. Sample DSMs for OMA I generated from MC-CNN+SGM trained using (a) satellite-training (the 2019 DFC benchmark) and (b) KITTI datasets.

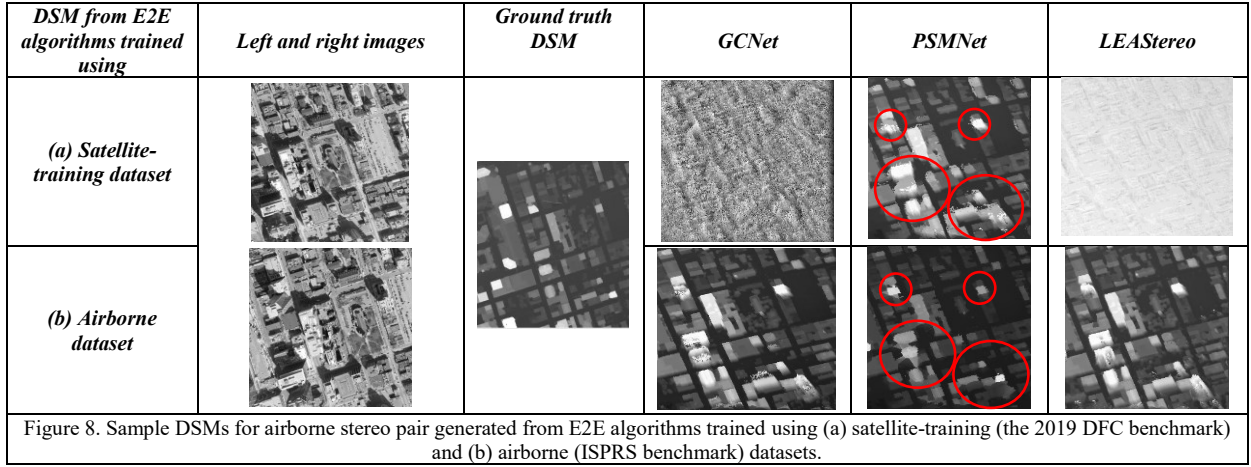

Figure 8. Sample DSMs for airborne stereo pair generated from E2E algorithms trained using (a) satellite-training (the 2019 DFC benchmark) and (b) airborne (ISPRS benchmark) datasets.

## VI. CONCLUSION

This work presents a comparative study of the traditional and DL stereo DIM algorithms, particularly for satellite datasets. We perform a comprehensive evaluation over a large volume of satellite data of 2,681 stereo pairs covering nine different regions around the world. We evaluate the stereo DIM algorithms on three bases, the overall performance, robustness to varying acquisition configurations, and generalization of the DL models. We study three classes of algorithms including: 1) traditional algorithms (e.g., Census+SGM), 2) DL learning-based algorithms (MC-CNN+SGM), and 3) E2E algorithms (e.g., GCNet, PSMNet, and LEAStereo). The results of the experiments suggest the following conclusions:

1) Overall, E2E methods outperform other types of stereo methods as they can achieve the lowest and most frequent minimum errors. They have also proven to have the highest completeness rates (of greater than 92%) of DSMs and the most consistent performance within individual test-sites. However, they can also achieve the highest errors (>10 meters) in some test-sites, possibly due to the lack of generalization. We found LEAStereo as the top-performing algorithm, as it achieved the lowest RMSE for half of the tested pairs and has the highest completeness rates (as seen in Section V-B). In addition, E2E methods proved to be robust with respect to the acquisition configurations, as compared with traditional methods, and this may be largely due to their ability to learn contextual correlations. However, their poor performance on some of the pairs shows that these types of methods lack generalization. Although they achieve the best results in some of the pairs, their average RMSE over all pairs being lower than Census+SGM, suggests that E2E methods are still unpredictable when processing a large number of datasets.

2) Census+SGM as a traditional method, to our surprise, outperforms MC-CNN in terms of the overall average error and the number of test-sites having the minimum average RMSE. This shows an opposite conclusion to the original work of MC-CNN, the reason for which might be due to the noisy training data projected from the LiDAR point clouds.

3) MC-CNN+SGM (similarity-based training) proved to be robust towards acquisition configurations of the stereo pair images. In addition, due to its nature of learning similarity, has indicated experimentally in our work that it has superior generalization capability, as long as the training datasets are noise-free.

Although a well-generalized stereo matcher with superior performance is ideal, our experiments show that in practice, the choice of stereo matcher can be case-specific. It is likely a well-trained E2E network produces superior quality results, provided that the training datasets are available, and the testing datasets are of the same type as the training datasets. The similarity learning-based method has good generalization capability, and if well-trained by datasets, may outperform traditional stereo matcher, although may not be as competitive as the E2E methods. Future research may improve the weaknesses of DL stereo DIM algorithms, and potentially consider domain adaptation methods to address some of the challenges. As for the quality of the training data, since it is imperative to the performance of DL DIM algorithms, we recommend further investigation on its impact to understand their implications and provide potential solutions.

## VII. ACKNOWLEDGMENT

The study is partially supported by the ONR grant (Award No. N000142012141). Hessah Albanwan is sponsored by Kuwait University. The author would like to thank John Hopkins University Applied Physics Lab to support the Imagery, the IARPA to organize the 3D challenge available that drives forth this work, the provision of the Downtown Toronto dataset by Optech Inc., First Base Solutions Inc., GeoICT Lab at York University, and ISPRS WG III/4.